\documentclass[letterpaper]{article} % DO NOT CHANGE THIS
\usepackage{aaai24}  % DO NOT CHANGE THIS
\usepackage{times}  % DO NOT CHANGE THIS
\usepackage{helvet}  % DO NOT CHANGE THIS
\usepackage{courier}  % DO NOT CHANGE THIS
\usepackage[hyphens]{url}  % DO NOT CHANGE THIS
\usepackage{graphicx} % DO NOT CHANGE THIS
\usepackage{natbib}  % DO NOT CHANGE THIS AND DO NOT ADD ANY OPTIONS TO IT
\usepackage{caption} % DO NOT CHANGE THIS AND DO NOT ADD ANY OPTIONS TO IT
\usepackage{algorithm}
\usepackage{algorithmic}

\usepackage{newfloat}
\usepackage{listings}
\usepackage{times}
\usepackage{epsfig}
\usepackage{graphicx}
\usepackage{amsmath}
\usepackage{amssymb}
\usepackage{graphicx}
\usepackage{amsmath}
\usepackage{amssymb}
\usepackage{booktabs}
\usepackage{nicefrac}
\usepackage{multicol}
\usepackage{multirow}
\usepackage{diagbox}
\usepackage{xcolor}

\usepackage{color, colortbl}
\definecolor{Gray}{gray}{.9}
\newcolumntype{g}{>{\columncolor{Gray}}c}
\usepackage{soul}
\def\eg{\emph{e.g. }} 
\def\ie{\emph{i.e. }} 
\def\etc{\emph{etc. }} 
\usepackage[capitalize]{cleveref}
\crefname{section}{Sec.}{Secs.}
\Crefname{section}{Section}{Sections}
\Crefname{table}{Table}{Tables}
\crefname{table}{Tab.}{Tabs.}

\newcommand\mypara[1]{\vspace{1mm}\noindent\textbf{#1}}
\def\btheta{\boldsymbol{\theta}}
\def\bX{\mathbf{X}}
\def\bY{\mathbf{Y}}

\DeclareCaptionStyle{ruled}{labelfont=normalfont,labelsep=colon,strut=off} % DO NOT CHANGE THIS
\floatstyle{ruled}
\newfloat{listing}{tb}{lst}{}
\floatname{listing}{Listing}
\title{SimCS: Simulation for Domain Incremental Online Continual Segmentation}
\author {
    Motasem Alfarra\textsuperscript{\rm 1,2},
    Zhipeng Cai\textsuperscript{\rm 1},
    Adel Bibi\textsuperscript{\rm 3},
    Bernard Ghanem\textsuperscript{\rm 2},
    Matthias Müller\textsuperscript{\rm 1}
}
\affiliations{
    \textsuperscript{\rm 1}Intel Labs\\
    \textsuperscript{\rm 2}King Abdullah University of Science and Technology (KAUST)\\
    \textsuperscript{\rm 3}University of Oxford\\
    motasem.alfarra@kaust.edu.sa
}

\usepackage{bibentry}
\begin{document}

\maketitle
\begin{abstract}
\label{abstract}

Continual Learning is a step towards lifelong intelligence where models continuously learn from recently collected data without forgetting previous knowledge. Existing continual learning approaches mostly focus on image classification in the class-incremental setup with clear task boundaries and unlimited computational budget. This work explores the problem of Online Domain-Incremental Continual Segmentation (ODICS), where the model is continually trained over batches of densely labeled images from different domains, with limited computation and no information about the task boundaries. ODICS arises in many practical applications. In autonomous driving, this may correspond to the realistic scenario of training a segmentation model over time on a sequence of cities. We analyze several existing continual learning methods and show that they perform poorly in this setting despite working well in class-incremental segmentation. We propose SimCS, a parameter-free method complementary to existing ones that uses simulated data to regularize continual learning. Experiments show that SimCS provides consistent improvements when combined with different CL methods. 

\end{abstract}

\section{Introduction}\label{introduction}

% \footnote{Correspondance to: motasem.alfarra@kaust.edu.sa.}
Supervised learning has been the go-to solution for many computer vision problems~\cite{he2016deep,ren2015faster}. The large scale of available labeled data has been the key factor for its success~\cite{clip}. However, in many settings the training data is not available all at once but generated sequentially over time. Moreover, the distribution of the training data may  vary gradually over time~\cite{cloc,lin2021clear}, \eg, images taken in winter with rain and snow versus images with clear skies taken in the summer. Naively applying supervised learning in such a setting suffers from \emph{catastrophic forgetting}~\cite{ewc}, \ie, training a model on new data of a different distribution worsens its performance on old data.
Continual learning (CL) attempts to address these issues by designing algorithms that operate on continuous data streams and efficiently adapt to new data while retaining previous knowledge. However, in the existing CL literature~\cite{li2017learning, agem}, methods are usually  evaluated only on restricted problems such as image classification with carefully crafted data streams that assume non-overlapping tasks, \eg, the class-incremental setting where each task corresponds to a 
fixed set of classes.

This work studies a more realistic problem of \textbf{O}nline \textbf{D}omain-\textbf{I}ncremental \textbf{C}ontinual Learning for Semantic \textbf{S}egmentation (ODICS). ODICS is essential for many applications where the perception system needs to be updated over time. In this setting, the model is trained with a \emph{limited computation and memory budget} at each time step on data sampled from a varying distribution~\cite{ghunaim2023real}. The variation comes from domain shifts, \eg, data coming from a different environment; the model has \emph{no information about the domain boundaries}. This setup mimics the practical scenario where labeled data from new scenes (weather conditions, cities, \etc) are generated continually over time, \eg, when developing a segmentation system for autonomous driving, last-mile delivery, or other robotics applications. The goal is to continually train the model (on the data center) with limited budget to enable frequent updates (\eg, once a day for self-driving cars) of deployed models.

Despite the importance of this problem, it has received little attention in recent years.
The few prior arts~\cite{douillard2021plop,maracani2021recall} for continual semantic segmentation study the problem under two unrealistic assumptions. First, it is assumed that the deployed model is aware of the \emph{domain boundaries}~\cite{dil}, \ie, the domain change, during both training and testing. While this simplifies the problem, domain boundaries are often not available in real-world applications as the transition between different domains is usually smooth or unknown. Second, the model is permitted to make any number of training iterations over current domain data, \ie, learning with  {unlimited computational budget}~\cite{douillard2021plop,dil,maracani2021recall}. This means that the model can pause the stream from revealing new data during training, while in realistic setups, streams continuously and uninterruptedly reveal new data and remain agnostic to the training status of the model~\cite{cloc,alfarra2023revisiting}.

\begin{figure*}
    \centering
    \includegraphics[width=0.9\linewidth]{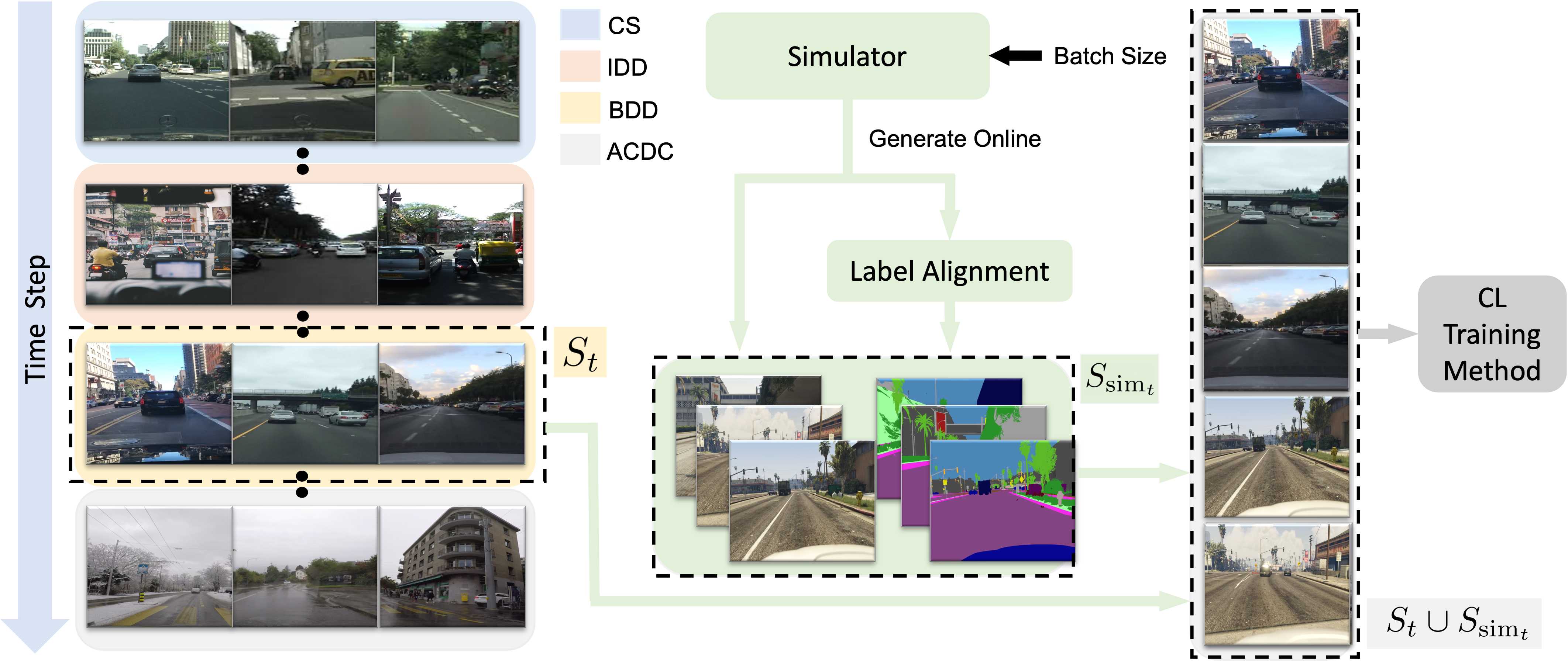}
    \caption{{Online Domain Incremental Continual Segmentation~(ODICS) with Simulated Data~(SimCS).}
    At each time step $t$, ODICS reveals a batch of labeled images $S_t$ with size $B_t$ from a certain domain, where different domains are presented sequentially to the model.
    SimCS generates a batch with size $B_t$ of simulated data $S_{\text{sim}_t}$ on the fly and aligns its label space to the real stream.
    The concatenated batch of real and simulated data is presented to the model to aid continual training and to mitigate forgetting previously learnt domains.
    }
    \label{fig:pull}
\end{figure*}
Moving closer to practical scenarios, we study the problem of online, \ie limited computational budget, domain-incremental continual learning for semantic segmentation. We propose a new benchmark using public datasets captured from different cities and different weather conditions and order them based on acquisition time.
We find that the domain shift in this benchmark is severe enough to cause forgetting on previously learned domains even without introducing new classes during training. We benchmark regularization-based methods~\cite{ewc} that are effective in mitigating forgetting in class-incremental continual segmentation~\cite{douillard2021plop} and show that they fail in ODICS. Meanwhile, although replay-based methods~\cite{chaudhry2019tiny} can effectively mitigate forgetting, 
they may not be feasible  due to privacy concerns, \eg, GDPR~\cite{gdpr}. This is particularly a concern when data is associated with different countries in which storing it for replay is not permissible. 
Thus, we propose SimCS that uses photo-realistic simulated data with ``free" dense labels which can be generated on the fly during ODICS~(see Figure~\ref{fig:pull}) to mitigate forgetting without violating privacy constraints. 

Our contributions are three-fold. \textbf{(1)} We propose online domain-incremental continual learning for semantic segmentation (ODICS) and construct a corresponding new benchmark evaluating several baselines from the literature. \textbf{(2)} We propose SimCS, a method that uses simulation as a continual learning regularizer. SimCS is parameter-free and \emph{orthogonal} to existing continual learning frameworks.  
We demonstrate its effectiveness by combining it with five different continual learning strategies showing performance improvements across the board. 
\textbf{(3)} We conduct a comprehensive analysis, showing that SimCS is robust to the choice of simulator, hyperparameters, 
and budget constraints in CL.

\section{Related Work}

\vspace{2pt}\textbf{Continual Learning.}
The main challenge in learning a sequence of tasks, classes, or domains from a continuous stream of data is catastrophic forgetting~\cite{wu2019large,rebuffi2017icarl}.
Existing methods can be broadly categorized into two groups, using either regularization or memory. \emph{Regularization-based methods} add regularization terms to the training objective without using previous data. The goal is to maintain parameters that are important to remember previous knowledge~\cite{ewc, mas, li2017learning}. Besides explicit regularization on model parameters~\cite{ewc, mas}, distilling the predictions of older models is also widely used~\cite{li2017learning}. \emph{Memory-based methods} store historical data it in a replay buffer~\cite{gem,gss}.
This  data is then used to regularize the gradient of the optimizer~\cite{agem} or directly mixed with new training data~\cite{chaudhry2019tiny}.
Most existing literature focuses on the class-incremental setup.
In this work, we evaluate the current progress in continual learning for semantic segmentation in the online domain-incremental setup.

\textbf{Continual Semantic Segmentation.}
Recently, class-incremental learning was extended from image classification to semantic segmentation.~\cite{michieli2019incremental,cermelli2020modeling}. 
This was done by presenting segmentation masks of the classes belonging to a given task while treating the remaining classes as the background~\cite{douillard2021plop,maracani2021recall}.
More closely related to our work, multi-domain incremental learning was analyzed in the semantic segmentation task~\cite{dil}. Despite the current progress, previous methods assume knowledge of task boundaries at test time and unlimited computational budget for training on each task~\cite{dil}. 
There are many practical applications like self-driving cars, where new data is generated constantly at a high data rate~\cite{cloc} and without clear distinction between different tasks~\cite{bang2022online}.
To better study these scenarios, we propose the setup of online domain incremental continual learning for semantic segmentation (ODICS), where a new batch of data arrives at each time step and the model is only allowed limited computation on each batch. 

\mypara{Simulators for Semantic Segmentation.}
Recent works have proposed several simulators that generate fully annotated data for ``free" such as CARLA~\cite{dosovitskiy2017carla} and VIPER~\cite{viper_simulator,richter2016playing}.
Such simulators play a key role for applications like autonomous driving~\cite{8959563} and visual navigation~\cite{li2020unsupervised}, where collecting and annotating data is expensive and time consuming.
Nonetheless, the use of simulated data in continual learning remains unexplored.
In this work, we leverage simulated data to reduce forgetting in continual learning.

\section{Online Domain-Incremental Continual Segmentation (ODICS)}\label{sec:benchmark}

In ODICS, a parametrized model $f(\cdot|\boldsymbol{\theta})$ that maps an image $\mathbf{X} \in \mathcal{X}$ to a per-pixel class prediction $\mathbf{Y} \in \mathcal{Y}$ is trained. At each time step $t \in \{1,2,3,...,\infty\}$, a batch of densely labeled images $S_t = \{\mathbf{X}_{i_t}, \mathbf{Y}_{i_t}\}_{i_t=1}^{B_t} \sim \mathcal{D}_t$ is revealed. Then, the model parameters $\boldsymbol{\theta}_t$ are updated using $S_t$ and a limited computation budget before $t+1$. Unlike in supervised learning where the domain $\mathcal{D}_t$ does not change, $\mathcal{D}_t$ may change drastically in ODICS during training. The goal of ODICS is to obtain the model parameters $\boldsymbol{\theta}_t$ that perform well on all previously seen domains, \ie, $\mathcal{D}_1$ to $\mathcal{D}_t$. There are two key concepts in ODICS: online and domain-incremental. \emph{Online} refers to the limited computation budget, \ie, we cannot train a model from scratch within each $t$. This is important for applications like autonomous driving where new data is continuously revealed over time. \emph{Domain-incremental} refers to the fact that the label space $\mathcal{Y}$ remains constant throughout training, \ie, only the distribution of $\mathbf{X}_{i_t}$ and the ratio of different classes in $\mathbf{Y}_{i_t}$ change over time. 
While ODICS is relevant for several applications, no benchmarks exist for this setup.
To this end, we propose a first attempt in constructing such a benchmark.
We focus on outdoor semantic segmentation in the context of self-driving cars where the domain shift can come from different weather conditions, cities, or cameras. 
To mimic practical scenarios, we construct the stream of data $\{S_t\}_{t=1}^{\infty}$ from multiple domains by composing four different standard benchmarks from the literature: CityScapes~(CS)~\cite{cordts2016cityscapes}, Indian Driving Dataset~(IDD)~\cite{varma2019idd}, Berkeley Driving Dataset~(BDD)~\cite{yu2018bdd100k}, and Adverse Weather Condition Dataset~(ACDC)~\cite{sakaridis2021acdc}; we treat each dataset as a different domain. Note that each dataset was collected in a different country. CS was collected in Germany, IDD in India, BDD in the United States, and ACDC in Switzerland mimicking the realistic scenario of deploying models in different locations.
This diversity introduces a notion of domains based on geographical location and weather conditions.
For example, CS contains images with clear weather conditions while ACDC has a variety of adverse weather conditions, \eg, fog and rain.
This adds another realistic aspect to our setup, since deployed models experience such adverse conditions when deployed throughout the year.

CS is used as reference for a consistent label space across domains, \ie, an identical set of classes.
We construct the stream by concatenating all domains based on the year the dataset was published, resulting in the following order: CS~(2016) - IDD~(2019) - BDD~(2020) - ACDC~(2021), which mimics the nature of continual learning where data generated earlier will be seen by the model first. For completeness, we analyze the use of different domain orders. 
As typical in CL, we evaluate the model trained on the stream on a held out test set from each domain.

\section{Methodology}

\subsection{Continual Learning Strategies}

We start with the scenario where at any time step $t$ the model cannot store, hence rehearse, any data from previous time steps (1 to $t-1$). 
This captures the realistic constraint where data is subject to privacy restrictions (\eg GDPR). 

The simplest baseline in this case is applying the same optimization strategy as in supervised learning at each time step, which we call \emph{naive training} in this paper. Specifically, given the training data $S_t = \{\mathbf{X}_{i_t}, \mathbf{Y}_{i_t}\}$ at time step $t$, we update the model by optimizing the following objective:
\begin{align}\label{eq:naive}
    \underset{\btheta_t}{\text{min}} \sum_{i_t}{\mathcal L(f(\bX_{i_t}|\btheta_t), \bY_{i_t})},
\end{align}
where $\mathcal L(\cdot)$ is the standard loss for semantic segmentation, \eg, cross entropy. In the online setting, we apply a limited number of (stochastic) gradient descent steps on $\btheta_t$.

\emph{Regularization-based methods}~\cite{ewc,li2017learning} are a family of continual learning methods extensively studied in image-classification. Instead of optimizing~\eqref{eq:naive}, these methods update the model by optimizing:
\begin{align}\label{eq:reg}
    \underset{\btheta_t}{\text{min}} \sum_{i_t}{\mathcal L(f(\bX_{i_t}|\btheta_t), \bY_{i_t})} + \lambda \mathcal L_{\text{reg}}(\btheta_t, \btheta_{t-k}, S_t),
\end{align}
where $\mathcal L_\text{reg}(\cdot)$ is the regularization term used to mitigate forgetting, which is algorithm-specific, and $\lambda$ is a coefficient that controls the regularization strength. Note that only data from the current step $S_t$ and possibly cached historical models $\btheta_{t-k}$ are used for regularization; no historical data is used for optimization.
We also consider relaxing the constraint on storing old data and complement our benchmark by comparing against \emph{replay-based methods}~\cite{chaudhry2019tiny}. In particular, we allow the model to store a few historical training samples in a small replay buffer. During training, 
we optimize:
\begin{align}\label{eq:replay}
    \underset{\btheta_t}{\text{min}} \sum_{i_t}{\mathcal L(f(\bX_{i_t}|\btheta_t), \bY_{i_t})} + \mathcal L_{\text{rep}}(\btheta_t, S_{\text{rep}_t}),
\end{align}
where $\mathcal L_{\text{rep}}(\btheta_t, S_{\text{rep}_t})$ computes the loss on a batch of data $S_{\text{rep}_t}$ sampled from the replay buffer at step $t$. In the simplest form~\cite{chaudhry2019tiny}, $\mathcal L_\text{rep}(\cdot)$ is the same as $\mathcal L(\cdot)$ but computed on $S_{\text{rep}_t}$. Despite its simplicity, this approach is effective in mitigating forgetting for image classification~\cite{prabhu2020gdumb}.

\subsection{Continual Learning with Simulation}\label{sec:approach}

In practice, naive training or regularization-based methods are often not effective enough for continual learning due to the strongly biased training data. Although replay-based methods are more effective, they are less practical under privacy or memory constraints. To address this problem, we take an orthogonal and unexplored path, which is using simulation data for continual learning.

Simulation techniques have achieved impressive advancements recently, especially for computer vision. For autonomous driving, state-of-the-art simulators~\cite{dosovitskiy2017carla,viper_simulator} can generate densely labeled images of simulated driving scenes on the fly. Using simulation for continual learning has several advantages. First, we can obtain an infinite amount of high-quality, diverse, and densely labeled images on the fly by running the simulator; we do not need to store a large amount of data in  memory. Second, since all data are synthetic, privacy constraints will not be violated.
Inspired by replay-based methods, we propose to use the loss on simulation data as a regularization for continual learning. As shown in Figure~\ref{fig:pull}, at each time step of ODICS, we first generate a batch of labeled simulation data $S_{\text{sim}_t}$ on the fly. Then, we update the model by optimizing the following objective:
\begin{align}\label{eq:sim}
    \underset{\btheta_t}{\text{min}} \sum_{i_t}{\mathcal L(f(\bX_{i_t}|\btheta_t), \bY_{i_t})} + \mathcal L_{\text{sim}}(\btheta_t, S_{\text{sim}_t}).
\end{align} 
We set $\mathcal L_{\text{sim}}(\btheta_t, S_{\text{sim}_t}) =  \sum_{{\bX_j, \bY_j} \in S_{\text{sim}_t}}{\mathcal L(f(\bX_j|\btheta_t), \bY_j)}$, \ie, we compute the same loss on both real and simulated data, and sum them together. 
While more complex strategies can be applied, we found that this simple approach is effective as later shown in the experiments. We call this method  Simulation for Continual Segmentation (\emph{SimCS}).
Note that the formulation in Eq.~\eqref{eq:sim} does not make any assumptions on $\mathcal L$.
That is, one can combine SimCS with regularization methods by replacing $\mathcal L$ with the combined loss in Eq.~\eqref{eq:reg}, or combine it with the replay approach in Eq.~\eqref{eq:replay}. This positions SimCS as an orthogonal approach to existing methods in the CL literature.
A few challenges need to be addressed to make SimCS general and effective.

\mypara{Robustness to simulators.} It is unclear whether our method can be robust to different simulators with different rendering quality, scene scale, and objects. To address this question, we use two different simulators, \ie, CARLA~\cite{dosovitskiy2017carla} and VIPER~\cite{viper_simulator}. 

\mypara{Label Space Alignment.} There are many options for defining class labels leading to misalignment between different simulators and real-world datasets. For example, CARLA has a single \textit{vehicle} class but separate classes for \textit{road line} and \textit{road}. Meanwhile, the real-world datasets have separate classes for \textit{cars} and \textit{trucks} but only a single class for \textit{road}.
Hence, we merge or relabel the segmentation masks generated by the simulator of choice to achieve the maximal overlap with the label space of the real data. Then, we drop all other labels as opposed to merging them into the background class. The full details of relabeling the simulated data can be found in the appendix. As shown in our experiments, though some of the real-world classes are missing in the simulated data due to non-overlapping label spaces, our approach is still effective. We expect that SimCS has further potential when applied to more advanced simulators.

\mypara{Data Quantity.} It is also not clear how much simulation data is needed for continual learning. Intuitively, using a large amount of simulated data could bias the model to only perform well on the simulated data, while using a small amount of data may only improve performance marginally. At each training iteration of SimCS, the batch of simulated data is generated by randomly setting simulator parameters, \eg camera position, weather, time and traffic conditions.
To study the impact of the amount of simulated data on performance, we explore varying the ratio between simulated and real data during training. In the main experiments, we set the sim-real ratio to 1, which provides a good trade-off between computation and performance improvement. 

\section{Experiments}~\label{sec:Exp}

\noindent\textbf{Experimental Setup.}
We construct our benchmark by concatenating four different datasets as domains, namely CS, IDD, BDD, and ACDC, as mentioned in Section~\ref{sec:benchmark}. 
Throughout, we use the term ``domain" and ``dataset" interchangeably. Following common practice in semantic segmentation~\cite{douillard2021plop}, we use 80\% of the publicly available data from each dataset for training and evaluate on the 20\% held out test set from each domain. 
We follow standard practice~\cite{douillard2021plop,maracani2021recall} in reporting the mean Intersection over Union~(mIOU) on the held out test set from each domain. 
During our experiments, at each time step $t$ of ODICS, the model is presented with a batch of real images $S_t$ of size 8, \ie $B_t = 8~\forall t$. Before the next time step $t+1$, the model is allowed to train on the batch using a fixed computational budget, measured by the number $N$
of forward and backward passes.
Unless stated otherwise, we set $N=4$ throughout our experiments\footnote{We found empirically that setting $N=4$ provides a good trade-off between preventing the model from under-fitting and significantly increasing the computation. In the appendix we provide results for different choices of $N$ with similar conclusions.}.
Once data from $S_{t+1}$ is revealed, the older batch $S_t$ becomes unavailable to the model unless replay is used. 
We evaluate all methods using the benchmark introduced in Section~\ref{sec:benchmark}.
In our experiments, we use the DeepLabV3 architecture~\cite{chen2017rethinking} pre-trained on ImageNet~\cite{deng2009imagenet} (unless otherwise stated in the pre-training experiments in Section~\ref{sec:pretraining}), following \cite{douillard2021plop}.
We utilized 2 NVIDIA V100 for each of our experiments.
Further details are provided in the appendix.

We analyze five different types of training strategies. The baseline is \emph{Naive Training (NT)}, \ie, optimizing Eq.~\eqref{eq:naive}. 
We also consider regularization-based (Eq.~\eqref{eq:reg}) and replay-based (Eq.~\eqref{eq:replay}) CL algorithms. 
For regularization-based methods, we consider \emph{Elastic Weight Consolidation~(EWC)}~\cite{ewc}, \emph{Memory Aware Synapses~(MAS)}~\cite{mas}, and \emph{Learning without Forgetting~(LwF)}~\cite{li2017learning}. 
We do not provide boundaries of dataset transitions during training; we make an exception for regularization-based methods, since this information is crucial to achieve reasonable performance according to our empirical results. 
For each considered regularizer, we set $\lambda = 0$ in Eq.~\eqref{eq:reg} when training on data from the first domain and $\lambda > 0$ for the other domains.
We report the best results for each regularizer cross-validated on different values of $\lambda$ leaving the result for all values of $\lambda$ to the appendix.
For replay-based methods, we apply Experience Replay~(ER)~\cite{chaudhry2019tiny} with a replay buffer size of 800 images (along with their dense labels), throughout this section and leave the ablations to the appendix.

\begin{table}[t]
    \centering
    \resizebox{\linewidth}{!}{
    \begin{tabular}{c|cccc|g}
    \toprule 
         \footnotesize{\backslashbox{Method}{Domain}} &  CS & IDD & BDD & ACDC &  mIOU\\
         \midrule
         NT $\qquad$&                   40.1  & 37.9  & 35.1 & 48.9 & 40.5 \\
          $\,\,\,\qquad$+ CARLA &           44.6  & 39.6 & 38.5  & 51.0 & 43.4 \\
          $\qquad$+ VIPER  &           45.4 & 43.8 & 40.0  & 50.4 & 44.9\\
         \midrule
          EWC$\qquad$ &                  41.5& 38.8&  35.9&  47.9& 41.0 \\ 
          $\,\,\,\qquad$+ CARLA &          45.3& 40.5& 38.8& 51.3& 44.0 \\ 
          $\qquad$+ VIPER &         45.1& 43.4& 40.9& 50.9& 45.1 \\ 
         \midrule
         MAS$\qquad$ &                  41.4& 37.1& 34.6& 48.2& 40.3 \\ 
          $\,\,\,\qquad$+ CARLA &          46.7& 41.3& 38.3& 50.2& 44.1 \\ 
          $\qquad$+ VIPER &          45.8& 43.5& 38.8& 49.1& 44.3 \\ 
         \midrule
         LwF$\qquad$ &              44.5& 41.9& 34.6& 46.3& 41.8 \\
          $\,\,\,\qquad$+ CARLA &         47.1& 44.3& 39.0& 48.5& 44.7 \\
          $\qquad$+ VIPER &        46.7& 46.7& 38.5& 47.9& 45.0 \\
         \midrule
        ER$\qquad$ &              47.4& 47.8& 40.9& 48.8& 46.2\\
          $\,\,\,\qquad$+ CARLA &         48.4& 48.5& 43.2 & 50.8&47.7 \\
          $\qquad$+ VIPER &        48.5& 50.0& 42.5& 52.0& 48.3\\
         \midrule
         Supervised &                   62.7& 63.6& 49.6& 62.0& 59.5\\
          $\,\,\,\qquad$+ CARLA & 62.8 & 63.7 & 49.7 & 62.9 & 59.7\\
           $\qquad$+ VIPER & 63.3 & 63.9 & 49.2 & 63.1 &59.8 \\
         \midrule
         \bottomrule
    \end{tabular}
    }\caption{{Performance Comparison under ODICS.} We report the mIOU~(\%) of a model trained on our benchmark and evaluated on each domain in the benchmark. 
    We also report the performance of SimCS-enhanced baselines by leveraging either CARLA or VIPER.
    All methods are trained with $N=4$ iterations for each received batch. The last row ``Supervised" represents the performance of a model trained on the entire stream for 30 epochs as a surrogate to upper bound performance.
    \emph{SimCS consistently improved the performance of all baselines on all observed domains.}
    }
    \label{tab:setup_1}
\end{table}

We explore simulated data generated from CARLA and VIPER~\cite{dosovitskiy2017carla,viper_simulator} with our SimCS approach.
We generate simulated data on the fly in the most realistic town 10 of CARLA by randomly setting the location and camera parameters. On the other hand, with VIPER, we sample (without replacement) from a large pool of the publicly available pre-generated simulated data, since the code to generate data on the fly is not available.
We relabel the segmentation masks of the simulated data to align with the labels of the real world following the procedure described in Sec.~\ref{sec:approach}.
This results in 13 and 15 out of 19 overlapping classes between the simulated and real data for CARLA and VIPER, respectively.

\subsection{Main Results}\label{sec:main_results}

We start by analyzing the performance of different CL training strategies in ODICS.
Table~\ref{tab:setup_1} summarizes the results of a model after being trained on our benchmark and evaluated on each observed domain, where the last column reports the mIOU across all domains.
The last row (Supervised) reports the performance of a model trained for 30 epochs using standard supervised learning on all data of the stream, representing a surrogate upper-bound
performance.

\begin{table}[t]
    \centering
    \resizebox{\linewidth}{!}{
    \begin{tabular}{c|cccc|g}
    \toprule 
         \footnotesize{\backslashbox{Method}{Domain}} &  CS & IDD & BDD & ACDC & mIOU\\
         \midrule
         NT $\qquad$&                   40.1  & 37.9  & 35.1 & 48.9 & 40.5 \\
         $\qquad$ + VIPER Pretrain & 40.2 & 40.4 & 36.5 & 51.9 & 42.3\\
         $\qquad$ + VIPER SimCS & 47.9 & 43.0 & 41.8 & 54.2 & 46.7 \\
         \midrule
         \bottomrule
    \end{tabular}
    }\caption{{Performance Comparison under VIPER Pretraining.}
    We compare the performance of NT when pretrained with VIPER (on top of ImageNet). We further boost NT + VIPER pretraining with SimCS (with VIPER) during continual learning.
    \emph{VIPER pretraining boosted the performance of both NT and NT+SimCS.}}
    \label{tab:viper_pretraining}
\end{table}

Unlike the class-incremental setup~\cite{douillard2021plop}, the simple NT in ODICS enjoys an on-par performance to all considered regularization-based methods.
For example, while MAS outperforms NT on earlier domains, \eg, CS, the overall performance degrades to 40.3\% compared to 40.5\% mIOU for NT. The most effective regularization-based method is LwF, which only outperforms NT by 1.3\%. This suggests that further work is needed to  develop regularization techniques for this more realistic domain-incremental setup. Meanwhile, rehearsing previously seen examples through ER consistently outperforms other baseline methods in all domains.
This conclusion is consistent with previous results in image classification~\cite{lin2021clear,gem,chaudhry2019tiny}, as storing real examples in a replay buffer provides a simple but  effective regularization for Continual Learning~(CL).

\begin{figure*}
    \centering
    \includegraphics[width=0.99\linewidth]{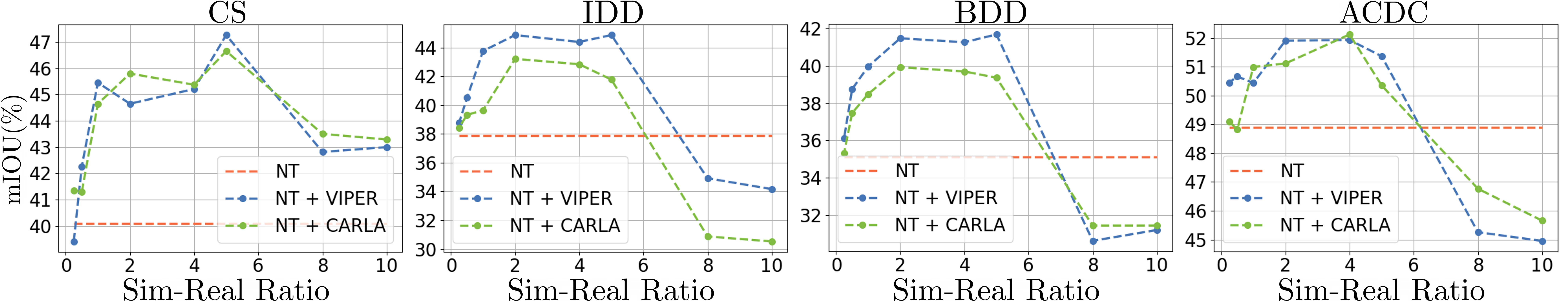}
    \caption{{Effect of Varying Sim-Real Ratio on the Performance Gain.}
    We analyze the effect of varying the ratio between simulation and real data from \{\nicefrac{1}{4}, \nicefrac{1}{2}, 1, 2, 4, 5, 8, 10\} on the performance gain for each observed domain. We find that SimCS provides notable performance improvement on a wide range of ratios ($\leq$ 5).
    % However, for very large Sim-Real ratios, the model becomes biased towards simulated data, thus, degrading the performance on real data.
    }
    \label{fig:sim_to_real_ratio}
\end{figure*}

\begin{figure*}
    \centering
    \includegraphics[width=0.99\linewidth]{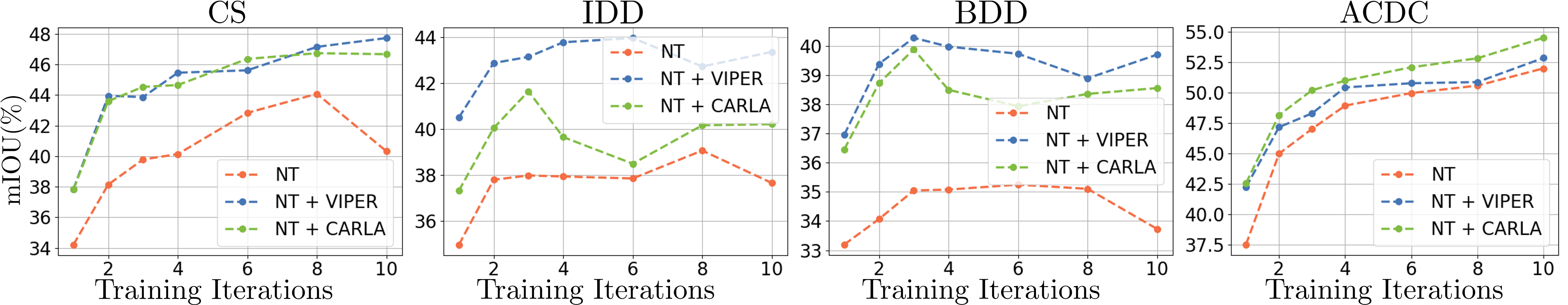}
    \caption{{Comparison under different computational budgets.}
    We allow NT, NT+CARLA, and NT+VIPER different computational budgets for training on each received batch from the stream, measured by the number of training iterations.
    We measure the performance on each observed domain when varying the budget to \{1, 2, 3, 4,6,8,10\} training iterations.
    % SimCS, provides consistent and large performance gains irrespective of the 
    % allowed computational budget and the choice of simulator.
    }
    \label{fig:computation_analysis}
\end{figure*}
Next, we analyze the effectiveness of including simulated data to all considered training strategies. To apply our approach on methods other than NT, we simply add $l_{\text{sim}}(\btheta_t, S_{\text{sim}_t})$ in Eq.~\ref{eq:sim} to the objective of each method. As shown in Table~\ref{tab:setup_1}, across \emph{all} domains and \emph{all} considered training schemes, SimCS provides consistent and significant performance improvements. 
For example, adding VIPER to the CL schemes reduces forgetting of NT and LwF on CS and IDD, respectively, by $\sim$ 5\% (from 40.1 to 45.4 and from 41.9 to 46.7).
Further, leveraging simulated data improves the strongest baseline (ER) by a notable 2\%. This result shows that simulation data can be leveraged as an effective regularizer for mitigating forgetting in CL. Moreover, different simulators provide different margins of improvements. For instance, while using either simulator (CARLA or VIPER) improves performance, simulated data generated from VIPER often produces larger gains. This observation can be attributed to several factors. For example, different simulators vary in photo-realism; in addition, their labels may be more or less aligned with real-word data labels.

\subsection{Pre-training on Simulated Data}\label{sec:pretraining}

In addition to using simulated data as a regularizer within the ODICS setup, we try incorporating it in pre-training, since it can be made available even before the start of the continual learning process. To that end, and before commencing with ODICS, we first fine-tune ImageNet pretrained models with data generated from VIPER.
We generate $17$K synthetic images, which is equal to the total number of real images presented in the continual setup, and train our model for 30 epochs on this simulated data.
We then perform ODICS and compare against NT and NT + VIPER, where in the latter VIPER is used as a regularizer during ODICS.

We report the results in Table~\ref{tab:viper_pretraining}.
We observe that pre-training on simulated data further improves the performance of a continual learner on all observed domains.
We report an improvement of $1.7\%$ on average across all observed domains when compared to ImageNet pre-training.
Although our model observed data generated from VIPER in the pre-training phase, including simulated data in the continual learning process further enhances the performance by 4\% on average.
It is worth mentioning that this boosted version of NT surpasses the performance of the best baseline ER by $0.5\%$ on average, without the need to store any new additional data from previous domains during continual learning.

\begin{figure*}
    \centering
    \includegraphics[width=0.99\textwidth]{
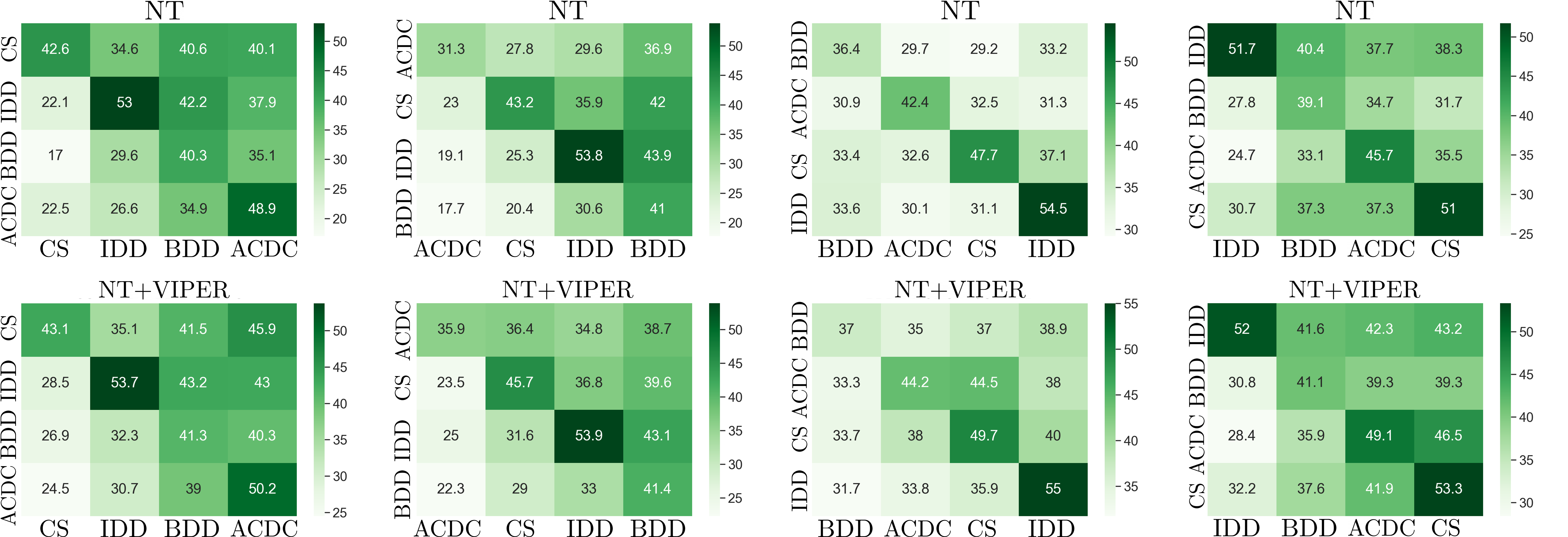}
    \caption{{Forward and backward transfer under different domain orders.}
    We analyze the forward and backward transfer during ODICS of both NT and NT+VIPER under different domain orders. The x-axis represents the observed domain within the stream while the y-axis shows the domain, on which we are evaluating the model. SimCS with VIPER improves both the forward (lower triangular) and backward (upper triangular) transfer in ODICS under different domain orders.
    }
    \label{fig:forgetting_analysis}
\end{figure*}

\subsection{Impact of the Amount of Simulation Data}\label{sec:ratio_sim_to_real}
In Sections~\ref{sec:main_results} and~\ref{sec:pretraining}, we used a 1:1 ratio between simulated and real data to form a mini-batch during continual learning.
We analyze the effect of varying this sim-real ratio on performance in Figure~\ref{fig:sim_to_real_ratio}, where we report the performance of NT, NT+CARLA, and NT+VIPER with ratios of $\{\nicefrac{1}{4}, \nicefrac{1}{2}, 1, 2, 4, 5, 10\}$. The results show that leveraging simulated data provides consistent performance improvements for a wide range of sim-real ratios. As long as this ratio was smaller than 5, \ie, we generate 5 batches from the simulator for each received batch from the real-world stream, our approach provides significant gains across all observed domains. 
However, for larger sim-real ratios, \eg 10, the training is biased towards simulated data and thus harms the performance on the real-world stream.
This is exemplified in Figure~\ref{fig:sim_to_real_ratio}, where 
SimCS with sim-real ratios $\geq$ 8 degrades the mIOU of NT on 3 out of 4 domains.
Further, VIPER outperforms CARLA across most sim-real ratios; this is consistent with previous observations in Table~\ref{tab:setup_1}.

\subsection{Impact of the Computational Budget}\label{sec:computational_budget}
In Section~\ref{sec:main_results}, all methods are given a fixed computational budget of $N=4$ forward and backward passes for each received batch.
In this section, we analyze the performance with different computational budgets. We conduct experiments with $N\in \{1, 2, 3, 4, 6, 8, 10\}$ for NT, NT+CARLA, and NT+VIPER, and report results on each observed domain in Figure~\ref{fig:computation_analysis}.
We observe that small computational budgets might result in an under-fitting model while larger budgets ($N=10$) cause the model to over-fit to the last domain, thus, increasing forgetting previous domains. Nonetheless, SimCS provides a stable performance gain across all considered budgets irrespective of the choice of the simulator.

Moreover, and in contrast to prior CL literature, we perform comparisons for when the computational budget is normalized for all methods, particularly when comparing NT against NT+VIPER. 
Since NT+VIPER uses a $1:1$ sim-real ratio in the batch, comparing it with NT using the same computational budget ($N=4$) might not be fair for NT. Effectively, NT+VIPER with $N=4$ is equivalent to $N=8$ due to the additional simulated data.
Our results in Figure~\ref{fig:computation_analysis} show that even when normalizing the computational budget,
SimCS still outperforms the baseline in ODICS.
For example, when NT+VIPER is allowed $N=4$ steps of computation, it achieves an mIOU of 43.8\% 
on IDD 
while NT with $N=8$ achieves 39.1\% 
on the same dataset. 

The interested reader may now wonder if the performance gains of SimCS simply come from allowing the model to train on more data. This is not the case. First, when allowing the model to train for a larger number of iterations, SimCS consistently improves performance as shown in Figure~\ref{fig:computation_analysis}, unlike the baseline. Second, only increasing the amount of data is not guaranteed to improve performance. The last row of Table 1 shows that simulated data does not help in the fully supervised setting.
Combining both observations shows that indeed SimCS improves performance by reducing forgetting. 
Finally, the cost of generating simulated data is negligible compared to the training cost.
For instance, the training time for a batch of 4 images for 4 iterations is 1.8 seconds, while generating a batch of 4 images from CARLA takes 0.18 seconds using the same hardware.
This adds a realistic advantage for SimCS, where simulated data can be generated on-the-fly during online learning.

\subsection{SimCS Improves Forward  Transfer}\label{sec:fwd_bwd}

Finally, we conduct a fine-grained analysis and
study the performance 
on all domains after training on every domain. 
Figure~\ref{fig:forgetting_analysis} summarizes this analysis, where the horizontal axis represents the last observed domain within the stream, while the vertical axis represents the domain we evaluate the model on. The last column corresponds to the results in Table \ref{tab:setup_1}, where we only report the performance of the final model after the last domain. 
At last,
we extend our analysis to different domain orders to include (ACDC-CS-IDD-BDD), (BDD-ACDC-CS-IDD), and (IDD-BDD-ACDC-CS).
Note that in each matrix, the performance difference ($r-d$) between a diagonal element $d$ and an element $r$ to the right of it in the same row reflects the forgetting (smaller is better). On the other hand, the performance difference $d-l$ between a diagonal element $d$ and an element $l$ to the left of it in the same row reflects the forward transfer.

First of all, including simulated data in the ODICS setup not only reduces forgetting, but also improves the forward transfer.
For example, including simulated data from VIPER boosts the forward transfer to ACDC in the (CS-IDD-BDD-ACDC) setup from 34.9\% to 39\% when trained on (CS-IDD-BDD).
We note that this result is not specific to the order at which the considered domains are presented. For instance, our approach improves the forward transfer from 20.4\% to 29\% on BDD when trained on (ACDC-CS) in the (ACDC-CS-IDD-BDD) setup.
Meanwhile, differently ordered streams result in larger variations in both forgetting and forward transfer.
For example, the performance on CS drops from 40.1\% to 36.9\% when changing the setup from (CS-IDD-BDD-ACDC) to (ACDC-CS-IDD-BDD). 
Furthermore, the performance on all domains (except IDD) drops significantly when IDD is the last domain.
Specifically, in the (BDD-ACDC-CS-IDD) setup, the forgetting on ACDC
is a significant 10.6\% mIOU.
This can be attributed to the distribution shift that IDD has compared to other domains.

\section{Conclusions}\label{sec:conclusion}

In this work, we investigated domain-incremental online continual learning for semantic segmentation. We identified the limitations of existing continual learning strategies and introduced SimCS, an orthogonal approach that utilizes simulated data generated on-the-fly to mitigate forgetting.
\section*{Acknowledgements}
This work was done during a research internship of the first author at Intel Labs.
This work was partially supported by the King Abdullah University of Science and Technology (KAUST) Office of Sponsored Research (OSR) under Award No. OSR-CRG2021-4648. 
We would like to thank Alejandro Pardo, Botos Csaba, and Shariq Bhat for the help and discussion.
Adel Bibi acknowledges the Amazon Research Awards funding.
\bibliography{aaai24}
\clearpage
\appendix
\clearpage
\section{CL with Simulation}

In Section~\ref{sec:approach}, we outlined the main details on how to include simulated data during ODICS.
Here, we elaborate on the implementation details such as the relabeling process.

\subsection{Relabeling CARLA}
\begin{table}[!htb]
    \caption{\textbf{Relabeling CARLA into CS label space.}}
    \centering
    \begin{tabular}{c|c|c}
    \toprule
    % \midrule
    Index Value & CARLA Label & CS Label\\
    \midrule
         0& Unlabeled& -  \\
         1& Building& Building \\
         2& Fence& Fence\\
         3& Other& -\\
         4& Pedestrian& Person\\
         5& Pole& Pole\\
         6& Road Line& Road\\
         7& Road& Road\\
         8& Side Walk& Side Walk\\
         9& Vegetation& Vegetation\\
         10& Vehicles& Car\\
         11& Wall& Wall\\
         12& Traffic Sign& Traffic Sign\\
         13& Sky& Sky\\
         14& Ground& -\\
         15& Bridge& -\\
         16& Rail Track& -\\
         17& Guard Rail& -\\
         18& Traffic Light& Traffic Light\\
         19& Static& -\\
         20& Dynamic& -\\
         21& Water& -\\
         22& Terrain& Terrain\\
        \midrule
         \bottomrule
    \end{tabular}
    \label{tab:carla_relabeling}
\end{table}
We deployed the stable v: 0.9.12 from CARLA to generate the simulated data for SimCS.
We employ our procedure in Section~\ref{sec:approach} while we relabel the generated data using Table~\ref{tab:carla_relabeling}.
We note here that all dropped labels (marked as '-') are not included in the loss calculation nor in model updates.

\subsection{Relabeling VIPER}
\begin{table}[!htb]
    \caption{\textbf{Relabeling VIPER into CS label space.}}
    \centering
    \begin{tabular}{c|c|c}
    \toprule
    % \midrule
    Index Value & VIPER Label & CS Label\\
    \midrule
1	& Ambiguous& - \\
2	& Sky& Sky \\
3	& Road& Road \\
4	& Side Walk& Side Walk \\
5	& Rail Track& - \\
6	& Terrain& Terrain \\
7	& Tree& - \\
8	& Vegetation& Vegetation \\
9	& Building& Building \\
10	& Infrastructure& - \\
11	& Fence& Fence \\
12	& Billboard& - \\
13	& Traffic Light& Traffic Light \\
14	& Traffic Sign& Traffic Sign \\
15	& Mobile barrier& - \\
16	& Fire Hydrant& - \\
17	& Chair& - \\
18	& Trash& - \\
19	& Trash Can& - \\
20	& Person& Person \\ 
21	& Animal& - \\
22	& Bicycle& - \\
23	& Motorcycle& Motorcycle \\
24	& Car& Car \\
25	& Van& Car \\
26	& Bus& Bus \\
27	& Truck& Truck \\
28	& Trailer& - \\
29	& Train& - \\
30	& Plane& - \\
31	& Boat& - \\
        \midrule
         \bottomrule
    \end{tabular}
    \label{tab:viper_relabeling}
\end{table}
We leveraged the available simulated data\footnote{VIPER data: \texttt{http://playing-for-benchmarks.org/}} released officially.
We relabel each pixel by using Table~\ref{tab:viper_relabeling}.
Similar to the previous section, we ignore all dropped labels from any loss calculations.

\section{Experiments}
Next, we present additional experimental ablations for ODICS and SimCS.
We first outline additional details about our experimental setup.
Then, we we present results with different computational budgets for all methods followed by ablating the effect of varying the memory size when using ER.
Further and for completeness, we report results for regularization based methods under different values of $\lambda$.

\subsection{Experimental Setup}
During our experiments we set the learning rate to $7\times10^{-3}$ throughout all of our experiments, following~\cite{douillard2021plop}. 
For regularization based methods, we set $\lambda=1, 10, 50$ for MAS, EWC, and LwF, respectively.
For Experience Replay method, we deployed First In First Out (FIFO) algorithm for updating our replay buffer.
That is, recently received examples from the stream will replace examples stored the the beginning of the buffer.
For all of our experiments, we utilized 2 NVIDIA V100 GPUs to run each experiment.

\subsection{Experimenting with Different N}
\begin{table}[t]
    \centering
    \caption{\textbf{Performance Comparison under ODICS.} We report the mIOU~(\%) of a model trained on our benchmark and evaluated on each domain in the benchmark. 
    We also report the performance of SimCS-enhanced baselines by leveraging either CARLA or VIPER.
    All methods are trained with $N=1$ iterations for each received batch. 
    \emph{SimCS consistently improved the performance of all baselines on all observed domains.}
    }
    \resizebox{\linewidth}{!}{
    \begin{tabular}{c|cccc|g}
    \toprule 
         \footnotesize{\backslashbox{Method}{Domain}} &  CS & IDD & BDD & ACDC &  mIOU\\
         \midrule
         NT $\qquad$&                    34.2 & 35.0& 33.2& 37.5& 35.0\\
          $\,\,\,\qquad$+ CARLA &           37.9& 37.8 & 36.5  & 42.5 & 38.7 \\
          $\qquad$+ VIPER  &            37.9& 40.5 & 37.0  & 42.4 &39.5 \\
         \midrule
          EWC$\qquad$ &   34.5& 35.7&  33.4& 38.0& 35.4             \\ %lambda 10
          $\,\,\,\qquad$+ CARLA &    37.4& 37.0& 36.7& 41.5&  38.2     \\ 
          $\qquad$+ VIPER &        37.9& 41.5& 37.5& 42.3& 39.8 \\ 
         \midrule
         MAS$\qquad$ &  33.9& 34.7& 32.5& 37.2&    34.6   \\ %lambda=1
          $\,\,\,\qquad$+ CARLA &  36.0&  36.2&  35.0&  41.2&   37.1   \\ 
          $\qquad$+ VIPER &        37.1& 39.9& 35.5& 39.3&   38.0 \\ 
         \midrule
         LwF$\qquad$ &   34.2& 36.2& 31.2& 35.0&  34.2  \\%lambda=50
          $\,\,\,\qquad$+ CARLA &      36.3& 40.9& 35.1& 38.7&  37.8  \\
          $\qquad$+ VIPER &         36.7& 42.5& 33.5& 38.3& 37.8\\
        %  \midrule
        % ER~\cite{chaudhry2019tiny}$\qquad$ &             \\
        %   $\,\,\,\qquad$+ CARLA &          \\
        %   $\qquad$+ VIPER &        \\
         \midrule
         \bottomrule
    \end{tabular}
    }
    \label{tab:N_1}
\end{table}

Throughout our experiments, we reported the results with a computational budget of $N=4$. For completeness, we report here the results with $N=1$ mimicking a fast stream setup where the learner is allowed to do only one training iteration on each received batch. Table~\ref{tab:N_1} summarizes the results. Similar to our earlier observations, SimCS provides consistent performance improvements in the case $N=4$. This demonstrates the effectiveness of SimCS on different streaming scenarios.
At last, we note that the performance of all methods with $N=1$ is significantly lower than with $N=4$. 
This can be attributed to the complexity of the task and the limited number of training data.

\subsection{Memory Size in ER}
\begin{table}[t]
    \centering
    \caption{\textbf{Performance Comparison under Different Replay Buffer Sizes.} We report the mIOU~(\%) of a model trained ER with different buffer size reported between parenthesis.
    We also report the performance of SimCS-enhanced baselines by leveraging VIPER.
    All methods are trained with $N=4$ iterations for each received batch. 
    We observe that larger buffer sizes improves the performance under the ODICS setups.
    Further, \emph{SimCS consistently improved the performance of all baselines on all observed domains.}
    }
    \resizebox{\linewidth}{!}{
    \begin{tabular}{c|cccc|g}
    \toprule 
         \footnotesize{\backslashbox{Method}{Domain}} &  CS & IDD & BDD & ACDC &  mIOU\\
         \midrule
         ER(200) $\qquad$&   46.9&  46.2& 40.9& 49.7& 45.9 \\
          $\qquad$+ VIPER  & 49.2&  48.9& 42.2& 51.4& 47.9 \\
         \midrule
         ER(800) $\qquad$&   47.4& 47.8& 40.9& 48.8& 46.2 \\
          $\qquad$+ VIPER  & 48.5& 50.0& 42.5& 52.0& 48.3 \\
         \midrule
        ER(1000)$\qquad$ &   48.6&  50.4& 42.2& 48.5& 47.4  \\
          $\qquad$+ VIPER &  49.2&  50.2& 43.5& 51.5& 48.6 \\ 
         \midrule
        ER(1200)$\qquad$ &       49.3&  50.2& 43.2& 50.3& 48.3 \\ 
          $\qquad$+ VIPER &      50.8&  51.7& 44.1& 50.3& 49.2  \\ 
         \midrule
         \bottomrule
    \end{tabular}
    }
    \label{tab:memory}
\end{table}
Next, we analyze the effect of the size of the replay buffer when deploying ER~\cite{chaudhry2019tiny}.
In particular, we experiment with a memory size of $\{200, 800, 1000, 1200\}$.
We report the results on Table~\ref{tab:memory} of the performance of ER under different buffer sizes with or without SimCS (using VIPER).
For this experiment, we set the number of training iterations to $N=4$ following our main setup in Seciton~\ref{sec:Exp}.

We observe that as the buffer size increases, the performance of the learner improves.
This is consistent with the earlier observations in the literature~\cite{chaudhry2019tiny} as larger buffer sizes allow the model to rehearse more diverse examples including several domains.
We note here that this comes at the expense of requiring large memory consumption in order to store more examples along with their segmentation masks.
Further, we find that SimCS provides a consistent performance improvement irrespective of the buffer size.
This shows another aspect of the robustness and how versatile SimCS under different learning algorithms for ODICS.

\subsection{Varying Regularization Importance}
\begin{table}[t]
    \centering
    \caption{\textbf{Performance Comparison under Different Regularization Importance.} We report the mIOU~(\%) of a model trained with different values of $\lambda$.
    All methods are trained with $N=4$ iterations for each received batch. 
    \emph{SimCS consistently improved the performance of all baselines on all observed domains.}
    }
    \resizebox{\linewidth}{!}{
    \begin{tabular}{c|cccc|g}
    \toprule 
         \footnotesize{\backslashbox{Method}{Domain}} &  CS & IDD & BDD & ACDC &  mIOU\\
         \midrule
         EWC$(\lambda=1)$ $\qquad$& 34.1 & 35.8& 33.1& 37.8& 27.7  \\
         EWC$(\lambda=10)$ $\qquad$& 41.5& 38.8&  35.9&  47.9& 41.0   \\
         \midrule
         MAS$(\lambda=1)$  $\qquad$&  41.4& 37.1& 34.6& 48.2& 40.3 \\
         MAS$(\lambda=10)$  $\qquad$&  36.7& 36.1& 31.4& 40.0&36.1  \\
         \midrule
        LwF$(\lambda=1)$  $\qquad$&  43.0& 38.9& 35.2& 47.9& 41.3  \\
        LwF$(\lambda=10)$ $\qquad$ & 43.1& 40.7& 36.3& 48.7& 42.2  \\
         \midrule
         \bottomrule
    \end{tabular}
    }
    \label{tab:lambda}
\end{table}
In Section~\ref{sec:Exp}, we reported the results for regularization based methods cross validated at the best $\lambda$ value. 
Here, we explore the effect of varying $\lambda$ on the performance of regularization based methods.
Table~\ref{tab:lambda} summarizes the results of EWC, MAS, and LwF for different regularization importance values $\lambda$.
We found that both EWC and MAS are more sensitive to variations of $\lambda$ as they deploy the regularization directly to network parameters.
However, we observe the distillation approach; LwF, is more robust to such variations making it more suitable to ODICS setup.

\begin{figure*}
    \centering
    \includegraphics[width=\linewidth]{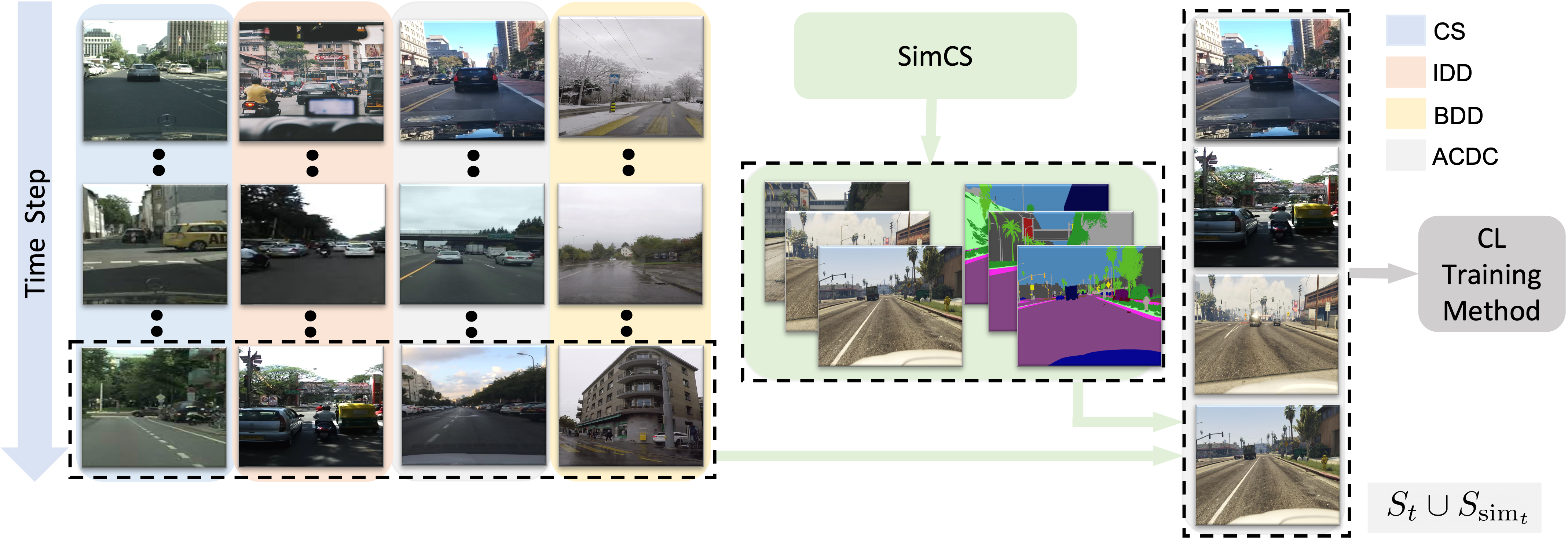}
    \caption{\textbf{Online Data Incremental Continual Semantic Segmentation with SimCS.} The learner receives a mixed batch from different domains. SimCS provides simulated data with aligned labels space to the real data.
    Note that different domains might present the learner with different amounts of data.
    }
    \label{fig:data_incremental_pipeline}
\end{figure*}
\subsection{Analysis on Different Domain Orders}
In Section~\ref{sec:fwd_bwd}, we analyzed the performance variations under different domain orders.
We also showed that SimCS improves, not only the backward transfer; \ie forgetting, but also the performance on unseen domains; \ie forward transfer.
Here we attempt at providing more insights on the reasoning for this behaviour.

Regarding improving the forward transfer, we believe that SimCS provides the learner with extra data diversity.
Meaning, simulated data could capture experiences that do not exist in earlier domains in the stream but might benefit the generalization to unseen domains.
Further, the distribution of labels in the generated simulation data could be closer to some real distributions than other real domains.
For example, one could match the number of car instances in a generated image to be closer to IDD (more vehicles) or change the weather conditions to match ACDC.
Note that SimCS, while generating the simulated data with randomly setting the simulation parameters, provided cross the board performance improvements on unseen domains as shown in Figure~\ref{fig:forgetting_analysis}.
We leave further experiments on better utilization of simulated data for CL for future work.

Regarding improving the backward transfer, we hypothesize the simulated data serve as a regularizer for not forgetting previously learnt domains.
That is, the distribution of simulated data does not exactly match any of the real-world domains, but might be bridging different real domains.
This is demonstrated on the robustness of SimCS to different domain orders in Figure~\ref{fig:forgetting_analysis} where performance improvements were shown in all considered scenarios.
Another evidence to this hypothesis, is the results reported in Table~\ref{tab:viper_pretraining} were SimCS improved the performance even when using VIPER in the pretraining stage.
That is, although the learner observed VIPER data, including simulated data in the continual learning process still provides significant performance enhancements.

\section{Online Data Incremental Setup}
\begin{table}[t]
    \centering
    \caption{\textbf{Performance Comparison Under Data Incremental Learning.} We report the mIOU of a model trained on a stream and evaluated on each of dataset in the stream.  }
    \resizebox{\linewidth}{!}{
    \begin{tabular}{c|cccc|g}
    \toprule 
         \footnotesize{\backslashbox{Method}{Dataset}} &  CS & IDD & BDD & ACDC & mIOU\\
         \midrule
         NT &                   42.3 & 47.7 & 38.9  & 39.7 & 42.1 \\
         $\qquad$ + CARLA &           43.4 & 49.8 & 39.2  & 41.4 & 43.4\\
         $\qquad$ + VIPER &           43.5 & 49.8 & 39.5  & 40.8 & 43.4\\
         \midrule   
         \bottomrule
    \end{tabular}
    }
    \label{tab:setup_2}
\end{table}
% \begin{table}[t]
%     \centering
%     \caption{\textbf{Performance Comparison Under Data Incremental Learning.} We report the mIOU of a model trained on a stream and evaluated on each of dataset in the stream.  }
%     \resizebox{\linewidth}{!}{
%     \begin{tabular}{c|cccc|g}
%     \toprule 
%          \footnotesize{\backslashbox{Method}{Dataset}} &  CS & IDD & BDD & ACDC & mIOU\\
%          \midrule
%          NT &                   51.1 & 52.9 & 43.0  & 49.4 &  \\
%          $\qquad$ + CARLA &           51.8 & 52.9 & 42.7  & 50.5 & \\
%          $\qquad$ + VIPER &           51.0 & 54.4 & 42.8  & 50.1 & \\
%          \midrule
%          \bottomrule
%     \end{tabular}
%     }
%     \label{tab:setup_2}
% \end{table}

At last, we present another realistic setup of continual learning for autonomous driving systems.
We consider the scenario where different systems exist at different locations and collect data simultaneously.
This results in a data incremental setup where the learner observes at each time step $t$ a set of images belonging to different domains.
However, different domains could reveal different amounts of data.
Figure~\ref{fig:data_incremental_pipeline} summarizes this setup with the inclusion of SimCS.

We conduct preliminary experiments with Naive Training~(NT) and report the results in Table~\ref{tab:setup_2} for NT, and NT+SimCS.
We follow similar experimental setup to the one in the main paper.
We observe that SimCS provides performance improvements under this setup, similar to ODICS.
We measure a performance improvement of $1.3\%$ when including simulated data into the training process.
We leave a further detailed analysis to this setup along with more experimental results to a future work.

\end{document}